\documentclass[journal]{IEEEtran} 
\usepackage[switch]{lineno} 

\usepackage{graphicx,color}
\usepackage{amsmath,amssymb,algorithm,xcolor}
\usepackage{float} 
\floatstyle{plaintop}
\restylefloat{table}
\usepackage{algpseudocode} 
\usepackage{array, multirow} 
\usepackage{ulem} 
\usepackage{comment} 
\usepackage{booktabs} 
\usepackage{epstopdf} 
\usepackage{url} 
\usepackage{subfig} 
\usepackage{placeins} 
\usepackage{tikz, pgfplots} 
\usepackage{enumitem}
\usepackage{soul}
\usepackage{cite} 
\newcolumntype{P}[1]{>{\centering\arraybackslash}p{#1}} 
\newcolumntype{L}{>{\raggedright\arraybackslash}m{3cm}}
\newcolumntype{R}{>{\raggedright\arraybackslash}m{3.5cm}}
\newcolumntype{Q}{>{\raggedright\arraybackslash}m{4cm}}
\newcolumntype{M}{>{\raggedright\arraybackslash}m{2.5cm}}
\newcolumntype{Z}{>{\raggedright\arraybackslash}m{6cm}}
\usepackage{fourier} 
\usepackage{array}
\usepackage{makecell}
\usepackage{dblfloatfix}
\begin{document}

\title{A Review of Prospects and Opportunities in Disassembly with Human-Robot Collaboration}

\author{Meng-Lun~Lee$^1$, Xiao Liang$^{2}$, Boyi Hu$^{3}$, Gulcan Onel$^{4}$, Sara Behdad$^5$, and Minghui~Zheng$^{1,*}$
	\thanks{$^1$Meng-Lun Lee and Minghui Zheng are with the Department of Mechanical and Aerospace Engineering, University at Buffalo, Buffalo, NY 14260, USA. E-mails: \{menglunl, mhzheng\}@buffalo.edu.}
	\thanks{$^2$Xiao Liang is with the Department of Civil, Structural and Environmental Engineering, University at Buffalo, Buffalo, NY 14260, USA. E-mail: \{liangx\}@buffalo.edu.}
	\thanks{$^3$Boyi Hu is with the Department of Industrial \& Systems Engineering , University of Florida, Gainesville, FL 32611, USA. E-mail: \{boyihu\}@ise.ufl.edu.}
	\thanks{$^4$Gulcan Onel is with the Department of Food and Resource Economics, University of Florida, Gainesville, FL 32611, USA. E-mail: \{gulcan.onel\}@ufl.edu.}
	\thanks{$^5$Sara Behdad is with the Department of Environmental Engineering Sciences, University of Florida, Gainesville, FL 32611, USA. E-mail: \{sarabehdad\}@ufl.edu.}
	\thanks{*Correspondence to Minghui Zheng.}
}

\maketitle    

\begin{abstract}
Product disassembly plays a crucial role in the recycling, remanufacturing, and reuse of end-of-use (EoU) products. However, the current manual disassembly process is inefficient due to the complexity and variation of EoU products. While fully automating disassembly is not economically viable given the intricate nature of the task, there is potential in using human-robot collaboration (HRC) to enhance disassembly operations. HRC combines the flexibility and problem-solving abilities of humans with the precise repetition and handling of unsafe tasks by robots. Nevertheless, numerous challenges persist in technology, human workers, and remanufacturing work, that require comprehensive multidisciplinary research to bridge critical gaps. These challenges have motivated the authors to provide a detailed discussion on the opportunities and obstacles associated with introducing HRC to disassembly. In this regard, the authors have conducted a thorough review of the recent progress in HRC disassembly and present the insights gained from this analysis from three distinct perspectives: technology, workers, and work.
\end{abstract}

\section{Introduction and Motivations} \label{sec:rv1} 
Environmental regulations, growing consumer demand for eco-friendly products, resource scarcity, and the potential profitability of salvaging operations have sparked a heightened interest in end-of-use (EoU) product recovery. This has prompted manufacturers to incorporate remanufacturing into their business models. Notably, even beyond environmental
\begin{figure}[!htbp]
	\centering
	\includegraphics[width=0.45\textwidth]{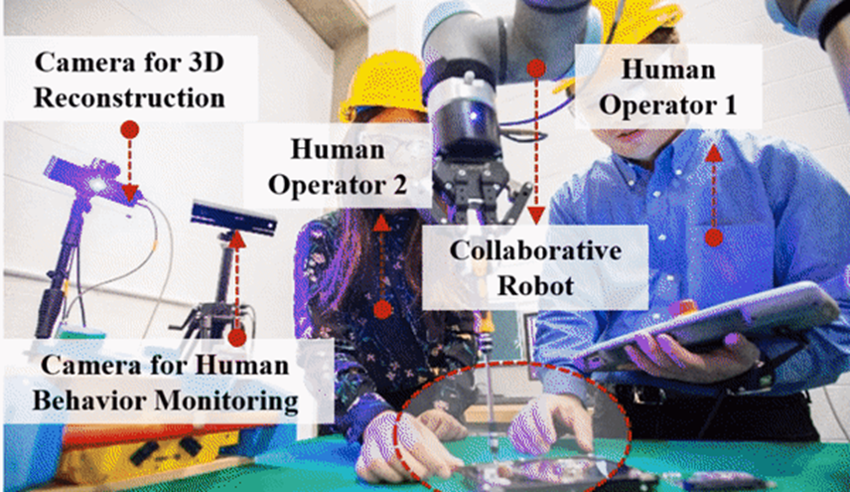}
	\caption*{Fig. 1: Human-Robot Collaborative Disassembly}	
	\label{fig:new1}
\end{figure}considerations, corporations have recognized the 
\begin{figure*}
    	\centering
	\includegraphics[width=0.9\textwidth]{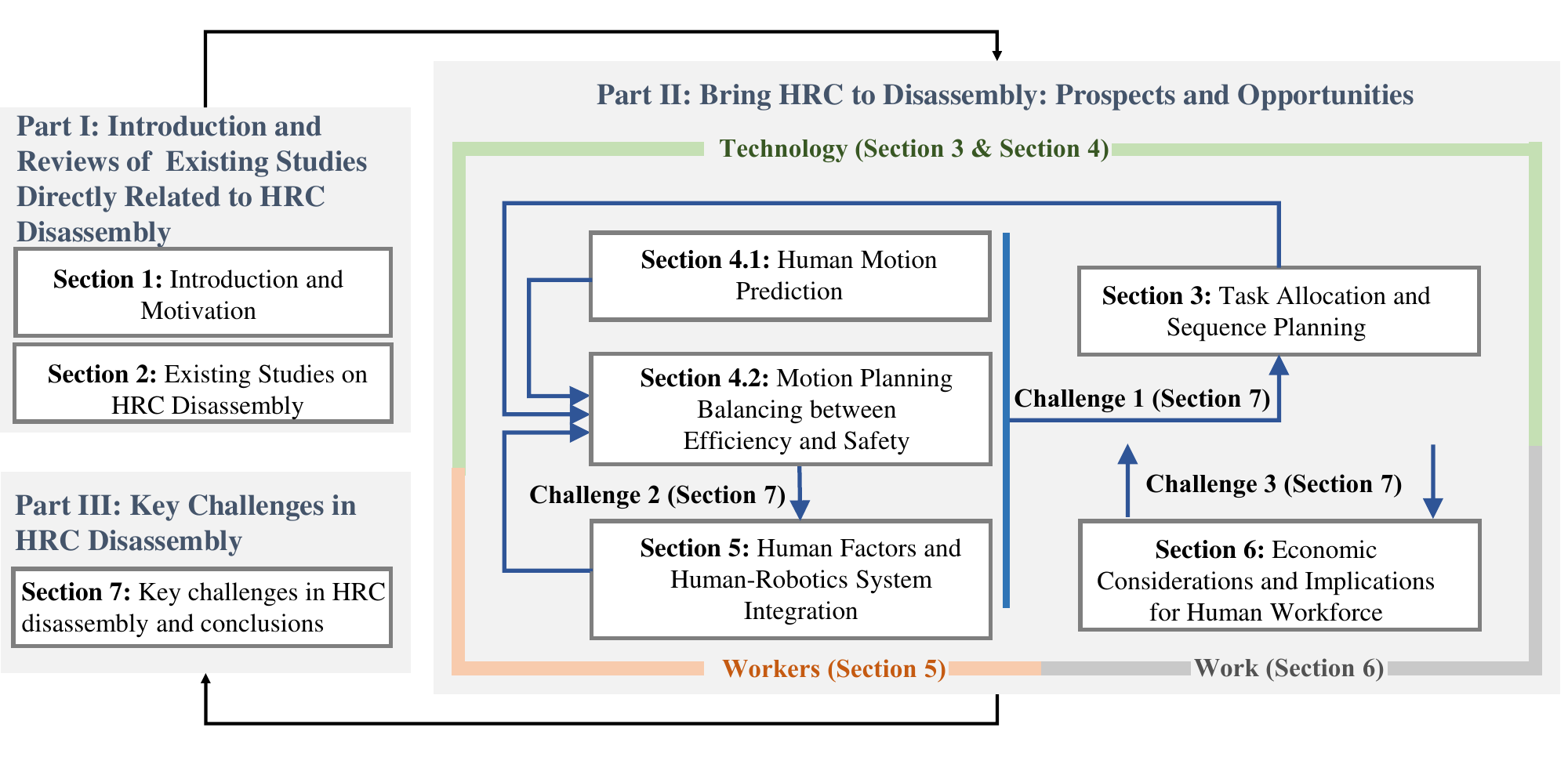}
	\caption*{Fig. 2: Paper Overview: This paper consists of 7 sections in total, which can be organized into three major parts. Sections 1 and 2 (Part 1) introduce the motivation for human-robot collaborative disassembly and review existing studies in this domain; Sections 3 to 6 (Part 2) focus on challenges from three perspectives, i.e., technology, workers, and work. In particular, Section 3 discusses the challenges that HRC brings to task allocation among humans and robots and to sequence planning of disassembly. Sections 4.1 and 4.2 discuss the challenges in planning robotic motion while balancing efficiency of robots and safety of humans. Section 5 discusses the challenges in human factors and human-robotics system integration. Section 6 discusses the economic considerations and implications for the human workforce. Section 7 (Part 3) summarizes the challenges of integrating all three perspectives (workers, work, and technology) for real remanufacturing systems.}	
	\label{fig:new2}
\end{figure*}economic value of what they once considered as trash. Industrial examples have demonstrated that remanufactured parts can be priced up to 50\% lower than new parts, showcasing the 
potential cost savings \cite{dornfeld2012green}. Implementing strategies such as optimizing the recycling process, embracing remanufacturing, improving disassembly techniques, and expanding repair and maintenance services all have positive impacts on businesses \cite{lund1996remanufacturing}. For manufacturers, remanufacturing their own products allows them to explore innovative business models, such as product-service systems and subscription-based models, while gaining greater control over the market by creating their own product ecosystems. From a socioeconomic perspective, semi-manual remanufacturing processes contribute to increased employment rates by generating new tasks and driving the need for expanded repair and remanufacturing efforts. The remanufacturing industry has shifted its focus from cost minimization to value creation for the broader social and economic systems. This paradigm shift necessitates interdisciplinary research and coordination across multiple fields, including economics, policy, occupational health, and engineering, among others.

Despite the inspiring vision and successful examples of remanufacturing efforts, there exist significant barriers to designing an effective remanufacturing system. Challenges such as the labor-intensive nature of disassembly \cite{esmaeilian2016evolution}, high labor costs \cite{lambert2007optimizing}, small lot sizes, uncertainty regarding the quality of incoming cores \cite{mashhadi2015uncertainty}, core unavailability, and the lack of automation \cite{giutini2003remanufacturing} hinder efficient and profitable remanufacturing. Remanufacturers often face difficulties in maintaining control over the supply chain, as they passively accept used products with uncertain quality, quantity, and conditions, which further complicates the remanufacturing and disassembly processes \cite{behdad2012end}. While manufacturers have made great strides in reducing manufacturing cycle times through the extensive use of robotic technologies in assembly processes, the reverse logistic aspect of the process still poses challenges. Disassembly remains predominantly labor-intensive, requiring direct contact with potentially harmful elements that can affect human health \cite{huo2007elevated,wang2018genomic,zhang2018maternal,cesta2016towards,li2019sequence}. Current manufacturing design guidelines, focused on optimizing assembly efficiency, often overlook disassembly considerations, resulting in suboptimal practices. The automation of disassembly remains an underdeveloped field.

Human-robot collaboration (HRC), employing collaborative robots, presents a promising solution for the labor-intensive disassembly process (Fig.~1). By leveraging the unique strengths and capacities of both humans and robots, the goal is to compensate for each other's weaknesses. Integrated efforts are required to improve the efficiency, safety, and sustainability of the disassembly line within the remanufacturing work domain.  Collaborative tasks such as disassembly, repair, and replacement can be carried out by robot manipulators and human workers, fostering a collaborative environment in remanufacturing factories (workplace).

Although there has been an increasing adoption of HRC in disassembly, many challenges in this domain have not been thoroughly identified, discussed, or studied. In the subsequent sections, we will conduct a comprehensive review of existing studies on HRC disassembly. Specifically, in Sections II and III, we will discuss several crucial perspectives that contribute to our understanding of HRC in disassembly. To shed light on the challenges and future directions of implementing HRC in remanufacturing, we will structure our discussions around four key perspectives: (1) Human-robot task allocation and distribution, (2) Robotic motion planning, (3) Human factors such as fatigue, and (4) Economics. These perspectives will provide valuable insights into the existing challenges and offer guidance for future research in the application of HRC to remanufacturing. By examining these four perspectives, we aim to identify the gaps that currently exist among technology, workers, and the work environment. Addressing these gaps is crucial to fully harnessing the potential of HRC in remanufacturing and ensuring its successful integration into industry practices. The overview and structure of this paper is provided in Fig.~2.

\section{Existing Studies on Human - Robot Collaborative Disassembly} \label{sec:HRCD}

\begin{table*}[!htbp]
	\centering
		\begin{tabular}{p{2cm}| p{1.5cm}|p{2cm}|R|L|p{1.8cm}}\toprule[1.5pt]
			\textbf{Collaboration Mode} & \textbf{Efficiency} &\textbf{Human \newline Involvement} &\textbf{Other advantages \& \newline challenges}&\textbf{Applications }&\textbf{References} \\\hline 
			Sequential disassembly  &Low  &Minimal& + Easy to implement \newline + Controlled process with less risk \newline - less adaptability to changing situations &Used for repetitive, well-defined disassembly tasks &\cite{fan2022multi, lee2020disassembly, lee2020real} \\\hline 
			Parallel \newline disassembly  & High &Significant& + Simultaneous part removal \newline - Coordination challenges in mixed teams& Used for resource-intensive disassembly, especially for complex products &\cite{huang2021experimental, parsa2021human, liu2019human, li2019sequence, xu2020disassembly, chatzikonstantinou2019new} \\\hline
			Collaborative disassembly &High  with optimized execution&Active\newline Collaboration&+ High flexibility in task allocation \newline + Leverages human-robot teaming \newline - Requires advanced safety measures \newline - Coordination challenges in mixed teams & Suitable for tasks that demand flexibility and human expertise &\cite{xu2021human, huang2020case, lee2022task, liu2019human2}\\
			\toprule[1.5pt]
		\end{tabular}
		\caption{Collaboration Modes}
		\label{tab:ch2_hrc_seq_par}
\end{table*} 

To gain insights into recent advancements in HRC disassembly, we conducted a meticulous review of relevant papers, with a particular emphasis on those published between 2010 and 2022. To facilitate our search, we utilized specific keywords such as ``human-robot" in conjunction with ``collaboration" or ``collaborative" along with ``disassembly," and explored renowned scientific databases, including IEEExplore.  and ScienceDirect. Subsequently, we manually screened the papers based on the following criteria:
\begin{itemize}
\raggedright
	\item  The involvement of at least one human operator and one robot working together in the disassembly process.
	\item  The papers presenting experiments or case studies involving the disassembly of real-world objects; hence, papers solely focused on simulations (e.g., virtual reality) were excluded.
\end{itemize}

Following a thorough examination of these papers, including two review papers \cite{poschmann2020disassembly, hjorth2022human}, it became apparent that studies directly related to HRC disassembly are limited in number. This section provides an extensive literature review, highlighting the significant contributions of these papers. The majority of the examined papers consider HRC in determining collaboration modes, defining disassembly objectives, as well as addressing safety and other relevant factors.

\subsection{Collaboration Modes} 
HRC makes the disassembly collaboration modes more complex than those without robots. Briefly, collaboration modes can be classified into three types: sequential \cite{fan2022multi, lee2020disassembly, lee2020real}, parallel \cite{huang2021experimental, parsa2021human, liu2019human, li2019sequence, xu2020disassembly, chatzikonstantinou2019new}, and collaborative \cite{xu2021human, huang2020case, lee2022task, liu2019human2}, as summarized in Table \ref{tab:ch2_hrc_seq_par}.

Sequential disassembly is a process in which parts are removed from EoU products one after the other, either by human operators or by robots. This mode is simple to implement but may not be the most efficient mode, as it can lead to bottlenecks in the disassembly process. On the contrary, parallel disassembly is a more efficient mode in which multiple components are removed simultaneously by human operators and robots. This mode can be more complex to implement, but it can significantly reduce disassembly time. Collaborative disassembly is a mode in which a human operator and a robot work together to perform a task. This mode can be the most efficient, but it requires careful planning and coordination between humans and robots. The choice of collaboration mode depends on the specific application. For example, for an EoU product with a large number of components, the parallel or the collaborative mode may be preferred because of their potential to reduce disassembly time. However, for a product with complex components, the sequential mode may be preferred because it allows for more precise control. 

\textbf{Sequential Disassembly.} Sequential disassembly is a mode of disassembly in which only one worker (the human operator or the robot) is assigned to a disassembly task at each step. This mode is simple to implement and ensures safe human operation, as the robot remains idling while the human operator performs a disassembly task, or the human operator maintains a safe distance and inspects the robot disassembling a component. Sequential disassembly has been studied in a number of studies. In \cite{lee2020disassembly}, a disassembly sequence planner was developed to sequentially assign disassembly tasks between one human operator and one robot. The planner was designed to avoid collisions between the human operator and the robot, and it was evaluated on a small EoU product. In \cite{fan2022multi}, a linear disassembly line with multiple workstations was studied. Each workstation consisted of a human operator and a robot, and the workers performed disassembly tasks one after the other. Additionally, in \cite{lee2020real, lee2022robot}, a receding-horizon sequence planner was proposed to distribute disassembly tasks to a human operator or a robot while factoring in real-time human motion. The planner traversed feasible task sequences and obtained the optimal one for the next three disassembly steps. This approach was shown to be effective in reducing the computational cost of planning disassembly sequences for EoU products with a large number of disassembly tasks. Overall, sequential disassembly is a simple and safe mode of disassembly that can be effective in some applications. However, it may not be the most efficient mode of disassembly, as it can lead to bottlenecks in the disassembly process.

\textbf{Parallel Disassembly.} Parallel disassembly is a mode of disassembly in which different tasks are assigned to human operators and robots simultaneously. This mode can significantly improve disassembly efficiency, as it allows workers to work on different tasks at the same time. There has been considerable research (Table \ref{tab:ch2_hrc_seq_par}) on parallel disassembly in the context of human-robot collaboration. For example, in \cite{huang2021experimental}, an HRC disassembly workstation with one human operator and multiple collaborative robots (cobots) was presented. The human operator and the robots could perform different disassembly tasks at the same time using force and position control. In \cite{liu2019human}, an HRC disassembly sequence planning for a diaphragm coupling was studied. In this work, distinctive disassembly tasks were assigned to a robot and a human operator simultaneously, but the working time for each task was assumed to be the same, regardless of the capability of the disassembly worker or the complexity of the disassembly task.  In \cite{parsa2021human}, a human-robot selective disassembly was proposed. In this approach, disassembly operations were carried out in parallel with tasks assigned to the human operator or the robot according to the complexity of each task. In \cite{li2019sequence} and \cite{xu2020disassembly}, HRC disassembly was considered as a parallel disassembly, so the optimal solutions to the HRC disassembly sequence were found by distributing tasks to a human operator and a robot simultaneously without violating precedence relationships of disassembly tasks. In \cite{chatzikonstantinou2019new}, components of EoU products were extracted through a series of operations, and the precedence relationships for the operations were assumed to be optional, which enabled human operators and robots to work on component extraction in parallel.

\begin{table*}[!htbp]
	\centering
		\begin{tabular}{R | p{2.6cm} | p{2.5cm} | R |p{3cm}}\toprule[1.5pt] 
			\textbf{Disassembly Objective} & \textbf{Sequential Mode} &\textbf{Parallel Mode}&\textbf{Collaborative Mode}&\textbf{References} \\\hline 
			Disassembly Time& Longer& Potentially Shorter&Optimizable for Efficiency&\cite{li2019sequence, chatzikonstantinou2019new, lee2022task} \\\hline
			Energy \newline Consumption & Lower & Higher&Potential for Optimization&\cite{fan2022multi, xu2020human, yin2022mixed} \\\hline
			Number of \newline  Workstations & Fewer &More&Variable, Optimizable&\cite{xu2020human, liu2019human2} \\\hline
			Complexity &Low & High&Adaptive and Varied&\cite{parsa2021human, li2019sequence, xu2020disassembly, lee2020disassembly, lee2020real} \\\hline
			Safety & Lower Risk&Increased Risk&Safety Protocols Needed& \multirow{2}{*}{\cite{li2019sequence, li2020unfastening, ding2019robotic}} \\\cline{1-4}
			Adaptability & Limited &Moderate&High Flexibility& \\
			\toprule[1.5pt]
		\end{tabular}
		\caption{Disassembly Objectives and Collaboration Modes}
		\label{tab:ch2_dsp_obj}
\end{table*}

\textbf{Collaborative Disassembly.} Collaborative disassembly is a mode of disassembly in which a human operator and a robot work together to perform the same task. This mode can be more efficient than parallel disassembly, as it allows the human operator and the robot to complement each other's strengths. For example, the human operator may be better at tasks that require dexterity and judgment, while the robot may be better at tasks that require strength and precision. By working together, the human operator and the robot can complete the task more quickly and efficiently. There has been some research on collaborative disassembly in the context of human-robot collaboration. In \cite{xu2021human} and \cite{liu2019human2}, human-robot collaborative disassembly task classification models were proposed. In these models, four types of worker groups were considered in distributing disassembly tasks: (i) only a human operator, (ii) only a robot, (iii) a human operator or a robot, and (iv) human-robot cooperation. In \cite{huang2020case}, the human operator and the robot worked together to separate parts in press-fitted components as HRC disassembly tasks. However, the study did not consider task sequence planning. Furthermore, a novel disassembly sequence planning (DSP) was conducted in \cite{lee2022task}.  This study proposed a DSP to obtain optimal disassembly sequence and distribute tasks among the robot, the human operator, and the collaborative human-robot team. The DSP considered disassembly time, transition between disassembly tasks, tool-changing time, and the limitation of the number of robots and human operators. The paper also studied the effect of assigning the human-robot team to the disassembly line and assumed that some disassembly tasks could be performed more efficiently by the human-robot team.

Parallel disassembly has been the focus of research in recent years. This mode allows human and robot workers to be assigned to different tasks simultaneously, which can significantly improve disassembly efficiency. However, sequential disassembly may be required in some cases, depending on the safety protocols of the disassembly process. Collaborative disassembly is the most important mode in HRC applications, but it is seldom discussed at the sequence planning level. To the best of our knowledge, \cite{lee2022task} is the only recent research that developed a disassembly sequence planner with a real-world case study that assigns one disassembly task to both the human operator and the robot, which is referred to as the collaborative disassembly in this paper.

\subsection{HRC Disassembly Objectives}

Once the disassembly modes are determined,  there are usually multiple objectives to be considered in planning a disassembly sequence. These objectives or considerations include but are not limited to, disassembly time \cite{lee2022task}, energy consumption \cite{yin2022mixed, xu2020human, fan2022multi}, number of workstations \cite{liu2019human2, xu2020human}, task complexity \cite{li2019sequence, parsa2021human, lee2020disassembly, xu2020disassembly} and others \cite{li2020unfastening, ding2019robotic, li2019sequence}. Table \ref{tab:ch2_dsp_obj} provides a brief summary on disassembly objectives and their correlation with disassembly modes.

\textbf{Disassembly Time.} The disassembly time is a significant objective in disassembly sequence planning, as minimizing the disassembly time also minimizes the labor cost. It is a common objective among a significant number of studies. For example, in \cite{li2019sequence}, the goal of the proposed HRC disassembly sequence was to minimize the total disassembly time, including the tool-changing time, the direction adjusting time, the moving time, and the waiting time. Similarly, in \cite{chatzikonstantinou2019new}, one of the disassembly objectives was to minimize the total disassembly time so that all operations in an HRC machine shop, consisting of multiple human operators, robots, and workstations, were performed in a timely manner. In \cite{lee2022task}, a proposed HRC disassembly sequence planning considered not only the total disassembly time of the human operator, the robot, and the collaborative human-robot team, but also the tool-changing time and the transition between different disassembly modules. In \cite{chu2023human} the disassembly completion time was the major objective for dismantling power batteries.

\textbf{Energy Consumption.} Energy consumption is an important objective in disassembly sequence planning, as minimizing energy consumption can maximize disassembly profit. This is because energy consumption is a major cost factor in disassembly, and it can also have a negative impact on the environment. Several studies have considered energy consumption as an objective in disassembly sequence planning. For example, in \cite{fan2022multi}, a human-robot collaborative disassembly line balancing problem (HRC-DLBP) was studied with the optimization objectives of minimizing energy usage and maximizing total profit. Similarly, in \cite{xu2020human}, an HRC-DLBP was explored to decrease the disassembly cost and increase the disassembly efficiency considering disassembly failure and energy consumption. In \cite{yin2022mixed}, an HRC-DLBP with multiple workstations was investigated to find the solutions for issues in optimization, including hazardous conditions, energy consumption, etc.

\textbf{Workstations.} Minimizing the number of workstations is an important objective in disassembly sequence planning, as it can improve work efficiency. This is because fewer workstations means that workers and robots can be more easily assigned to tasks, and it also means that there is less space required for the disassembly line. One of the disassembly objectives in \cite{xu2020human} for a simplified hammer drill disassembly was to minimize the number of workstations. Each workstation could perform one disassembly task assigned to either the human operator or the robot. In \cite{liu2019human2}, a set of HRC models was developed for DLBP with the goal of achieving several optimization objectives, including minimizing the number of disassembly workstations, balancing the workload, and trying to maintain the idle time consistency of each workstation.

\textbf{Disassembly Complexity.} Disassembly complexity is an important factor to consider in disassembly sequence planning, as it can affect the time, cost, and safety of the disassembly process. In \cite{parsa2021human}, a quantitative scoring algorithm was proposed to distribute disassembly tasks to the robot and the human operator according to the complexity levels as the disassemblability parameters for each disassembly task. In \cite{li2019sequence}, the complexity and uncertainty level of the disassembly tasks were considered in selecting the workers between the human operator and the robot in DSP, as the human operator could flexibly handle complex disassembly tasks and the robots can perform tasks repeatedly with high precision. In \cite{xu2020disassembly}, the concept of disassembly attributes was introduced to solve the HRC-DSP problem. The attributes included moving complexity, workload, hazardous level, etc. In \cite{lee2020disassembly} and \cite{lee2020real}, the geometric complexity and the labor efforts of the to-be-disassembled components were considered as the disassembly cost in finding the HRC optimal disassembly sequence. 

\begin{table*}[!htbp] 
	\centering
	\begin{tabular}{M |Z| Q}\toprule[1.5pt] 
			\textbf{Safety Strategy} &\textbf{Methods}& \textbf{References} \\\hline 
			Motion planning  & Prevent collision via planning robot motion in real time&\cite{huang2021experimental, liu2019human, xu2021human, huang2020case, li2019sequence, liu2019human2, corrales2012cooperative, gerbers2018safe} \\\hline
			Task planning  & Develop safety measures by considering factors such as
			hazardous components and working distances and assign tasks accordingly&\cite{lee2022task, lee2020disassembly, lee2020real} \\
			\toprule[1.5pt]
		\end{tabular}
		\caption{Disassembly Safety Strategy}
		\label{tab:ch2_seq_mot_tas}
\end{table*}

\textbf{Other Considerations.} Other disassembly objectives aimed to address specific problems within the disassembly process by utilizing HRC, without involving HRC disassembly sequence planning. In one example, the study by \cite{li2019sequence} incorporated a human fatigue factor in the HRC-DSP to prevent work deficiency resulting from continuous manual labor. Another study by \cite{li2020unfastening} presented an automated unfastening technique for threaded hexagon-headed screws, eliminating the need for human assistance by accurately locating the screws' position and centerline. \cite{ding2019robotic} focused on transferring valuable information from human operators to a robotic knowledge graph system to enhance disassembly efficiency in HRC disassembly tasks. Several studies have integrated multiple factors simultaneously. For instance, Liao et al. \cite{liao2023optimization} developed a framework that considered disassembly time, complexity, and operator safety, using a multi-attribute utility function to determine the optimal disassembly sequence and allocate work between humans and robots.  Guo et al.\cite{guo2023human} considered disassembly time, cost, and part condition/failure status to determine the optimal sequence for human-robot collaboration. Belhadj et al. \cite{belhadj2022product} generated disassembly plans by minimizing changes in dismantling directions and tools. It is important to note that, in addition to the aforementioned objectives, operator safety and well-being are of significant importance, and a dedicated section will thoroughly discuss these aspects.

Our analysis of the collected papers reveals that task complexity is the most frequently discussed objective in the disassembly of EoU products. The existing research on HRC-DSP often considers multiple factors when optimizing the disassembly process. Consequently, the specific disassembly objectives and the available data obtained from the disassembly process influence the parameterization of the disassembly cost, ultimately leading to the generation of an optimal HRC disassembly sequence.

\subsection{Consideration of Human}

Human operators play a critical role as partners to robots in the context of HRC in disassembly processes. Ensuring human safety is a primary focus in existing studies, with multiple methods employed to guarantee the well-being of human operators. These methods can generally be categorized into two levels: decision level (task planning) and robotic planning, as summarized in Table \ref{tab:ch2_seq_mot_tas}. At the task planning level, considerations are given to the hazardous conditions of components and the working distance between human operators and robots. At the robotic planning level, motion planning algorithms are integrated to prevent collisions with human operators.

\textbf{Motion Planning.}  In the existing literature, the safety strategy predominantly revolves around maintaining a safe distance between human operators and robots. If the distance falls below the specified safety requirements, the robot will slow down or come to a complete stop to ensure a safe workspace. However, this cautious approach can affect the efficiency of collaborative disassembly. For instance, Huang et al. \cite{huang2021experimental} implemented force and position control, along with active compliance control, in multiple collaborative robots equipped with torque sensors. They used a laser scanner in the proposed disassembly cell to achieve safe human-robot interaction and enable complex disassembly operations. Liu et al. \cite{liu2019human} proposed a safety assessment system utilizing a human skeleton point cloud model to calculate the minimum safe distance. Xu et al. \cite{xu2021human} considered the distance between human operators and robots for different disassembly tasks to ensure human safety, with the robot's speed adjusted based on this distance. Huang et al. \cite{huang2020case} implemented active compliance control in the robot to safely carry out complex disassembly operations alongside the human operator. Li et al. \cite{li2019sequence} introduced a human fatigue model for HRC disassembly workspaces, considering the accumulation rate of human fatigue to prevent workplace hazards. Liu et al. \cite{liu2019human2} presented a human-robot collaborative safety strategy combined with a line balancing optimization solution, wherein the robot adjusted its speed based on its distance to the human operator during each disassembly task. At the motion planning level, Corrales et al. \cite{corrales2012cooperative} deployed two types of safety strategies to track the distance between the human operator and the robot, avoiding collisions through complete stops and move-away actions. Gerbers et al. \cite{gerbers2018safe} presented a collaborative workstation for automated unscrewing, aiming to prevent human workers from being exposed to toxic materials during the disassembly of EoU electric vehicle batteries.

\textbf{Task Planning.} Considering potential hazardous materials or unsafe work environments, certain disassembly tasks should be allocated to the robot rather than the human operator to ensure safety. Therefore, safe conditions need to be taken into account when distributing tasks in HRC disassembly. For example, Lee et al. \cite{lee2022task} achieved safe HRC disassembly sequence planning by considering the safe conditions of components and the distances between disassembly tasks. In both Lee et al.'s works \cite{lee2020disassembly, lee2020real}, the safe condition of the components to be disassembled was considered during the sequence planning stage. This approach ensured that unsafe tasks were not assigned to the human operator, thus maintaining human safety.

\section{Bringing HRC to Disassembly: Task Allocation and Sequence Planning (Technology)} \label{sec:disassembly}

When introducing robots, one of the key challenges is determining the optimal disassembly sequence and effectively distributing and planning the disassembly tasks among humans and robots. Traditional DSP is often formulated as an NP-hard (non-deterministic polynomial-time hardness) optimization problem \cite{lee2022task, meng2016improved, bahubalendruni2021disassembly}, but it does not explicitly consider the involvement of robots and HRC. Several comprehensive review papers have been published on DSP from 2010 to 2018. The first one \cite{ilgin2010environmentally} reviewed over 500 papers from 1977 to 2010 with more than 80 articles relevant to DSP. The second \cite{ilgin2015use} collected about 200 papers up to the year 2015 with nearly 40 articles associated with DSP. Noting that both the two review papers \cite{ilgin2010environmentally, ilgin2015use} examined the state-of-the-art of designing a product considering its life cycle, including the recovery from its EoU stage. The third \cite{guo2020disassembly} surveyed 137 papers up to the year 2021 and classified DSP into four perspectives without HRC: modeling, mathematical programming, artificial intelligence (AI) techniques and uncertainty analysis in the disassembly process. The fourth one \cite{zhou2019disassembly} investigated about 150 papers from 1998 to 2018 examining DSP in the aspects of disassembly modes, disassembly modeling and planning methods. It is worth noting that HRC was not explicitly considered in those review papers mentioned above. 

\begin{table*}[!htbp] 
	\centering
		\begin{tabular}{M|Z|Q}\toprule[1.5pt]
			\textbf{Disassembly Mode} & Main Applications & \textbf{References} \\\hline
			Complete disassembly        &End-of-life product recycling, material separation, proper disposal of hazardous waste, etc.& \cite{yeh2011optimization, xia2016service, jin2014solution, alshibli2016disassembly, zhao2014fuzzy, tian2013chance} \\\hline
			Selective or target disassembly           &Component replacement, repair, product upgrade, customization, etc.& \cite{luo2012disassembly, smith2016partial, xia2014q, zhong2011disassembly, behdad2010simultaneous} \\
			\toprule[1.5pt]
		\end{tabular}
		\caption{Disassembly Modes}
		\label{tab:ch1_dsp_modes}
\end{table*}

In the following paragraphs, we will discuss several key principles of DSP and the challenges of bringing HRC into those principles: (i) disassembly modes (complete disassembly v.s. selective disassembly), (ii) disassembly modeling for generating feasible disassembly sequences, and (iii) optimization methods for obtaining optimal disassembly sequences.

\textbf{Disassembly Modes.} Determining the disassembly mode is the first step in DSP. The existing disassembly modes can be classified as total disassembly and selective disassembly \cite{luo2012disassembly}. Literally, total disassembly \cite{yeh2011optimization, xia2016service, jin2014solution, alshibli2016disassembly, zhao2014fuzzy, tian2013chance} involves dismantling an EoU product into individual components, which is expensive and impractical. In contrast, selective disassembly only removes valuable components from the EoU products \cite{luo2012disassembly, smith2016partial, xia2014q, zhong2011disassembly, behdad2010simultaneous}. The above disassembly modes and references are listed in Table \ref{tab:ch1_dsp_modes}. 

Due to significant uncertainty in remanufacturing systems, it is important that HRC systems are equipped with proper detection technologies to evaluate the potential value of recovering components before targeting to dismantle them. This is particularly important for selective disassembly. Recent studies have paid attention to this issue by developing disassembly metrics. For example,  \cite{yu2022disassembly} proposed an approach that involves constructing an ontology to describe component information and assembly relations, formulating destructive rules for guided disassembly, and generating feasible planning schemes. While the focus of the study was not on HRC, the proposed approach provides a foundation for automating the disassembly process toward maximizing the value retention of EoU devices. In \cite{ali2022quantitative}, a method was proposed for quantitatively evaluating the disassemblability of products. This method considered factors such as the quality of returns, product design characteristics, and process technology requirements. Such approaches should be extended to the HRC field to incorporate the limitations of both humans and robots when determining disassembleability.

\textbf{Modeling of Feasible Disassembly Sequences.} To plan a disassembly sequence effectively, it is crucial to consider various attributes of the components in the EoU product, such as the precedence relationships among the to-be-disassembled components, the geometric complexity in removing the parts, the presence of hazardous conditions of the disassembly, and the tools required for the disassembly operations. These factors could significantly impact the process of finding the desired disassembly sequence. There are multiple modeling methods in the existing literature: (1) Graph-based modeling is the one that uses undirected or directed graphs \cite{behdad2010disassembly, behdad2012disassembly, kuo2010waste, song2010product} or AND/OR graphs \cite{min2010mechanical, min2010research, tseng2010integrated, xia2019disassembly, guo2017dual, mutlu2021memetic} to represent the disassembly orders between different subassemblies or individual components; (2) Petri-net represents the structural relationship of the components, as studied in \cite{zhao2014fuzzy, kuo2010waste, giudice2010disassembly, kuo2013waste, guo2015disassembly}; (3) Matrix-based methods use the matrix form, known as the precedence matrix, the connection matrix or the constraint matrix, to describe the precedence relationships among to-be-disassembled components \cite{smith2011rule, smith2012multiple}. In addition, there are also researchers proposing different modeling methods to solve the DSP problems, including selective DSP \cite{peng2011selective, elsayed2012robotic, han2013mathematical, mitrouchev2015selective}, branch-and-bound based algorithm \cite{zhang2010product}, disassembly information modeling \cite{zhu2013disassembly} and fuzzy-rough set \cite{zhang2014parallel}. The above disassembly modeling methods and references are listed in Table \ref{tab:ch1_dsp_model}.

\begin{table}[!htbp]  
	\centering
	\begin{tabular}{p{4cm} | p{3.5cm}}\toprule[1.5pt]
		\textbf{Modeling Method} & \textbf{References} \\\hline
		Graph-based modeling         & \cite{behdad2010disassembly, behdad2012disassembly, kuo2010waste, song2010product, min2010mechanical, min2010research, tseng2010integrated, xia2019disassembly, guo2017dual, mutlu2021memetic, ma2011disassembly} \\\hline
		Petri-net modeling           & \cite{zhao2014fuzzy, kuo2010waste, giudice2010disassembly, kuo2013waste, guo2015disassembly} \\\hline
		Matrix-based modeling        &  \cite{smith2011rule, smith2012multiple} \\\hline
		Other modeling methods       &  \cite{zhang2010product, peng2011selective, elsayed2012robotic, han2013mathematical, zhu2013disassembly, zhang2014parallel, mitrouchev2015selective}\\
		\toprule[1.5pt]
	\end{tabular} 
	\caption{Disassembly Modeling Methods} 
	\label{tab:ch1_dsp_model} 
\end{table} 

While an extensive stream of literature exists on identifying feasible disassembly sequences, most of them assume manual disassembly. However, in the context of HRC, it becomes essential to develop approaches that can generate feasible sequences while considering the capabilities and limitations of both humans and robots. Along this line, studies toward automating the identification of feasible sequences would be beneficial as they can be augmented by HRC requirements. For example, \cite{prioli2022disassembly} proposed a process to extract geometrically feasible disassembly sequences from CAD assembly files. The feasibility of disassembly was assessed using a precedence matrix that indicated the order in which components can be safely removed.  \cite{upadhyay20233d} utilized 3D data from CAD assembly models to generate viable disassembly sequences by using Graph-based learning to process the graph representation of the CAD models.   

\textbf{Finding the Optimal Sequence and Task Allocation.} The DSP problem has been tackled by many optimization algorithms. An optimization objective in DSP is to conduct a disassembly sequence in an efficient way, which can be classified into nature-inspired algorithms, linear programming methods, and other optimization methods. Nature-inspired algorithms can be subclassified into the following groups: genetic algorithms \cite{go2012genetically, kheder2015disassembly, kucukkoc2020balancing}, artificial bee colony \cite{percoco2013preliminary, liu2018robotic, liu2020collaborative, wang2021discrete, kalayci2013artificial, ccil2022two}, ant colony \cite{luo2016integrated, malik2010performance}, particle swarm optimization \cite{zhong2011disassembly, yeh2012simplified, kalayci2013particle}, and scatter search \cite{li2013selective, kuo2013waste, guo2019lexicographic, guo2017dual}. Linear programming methods \cite{zhu2013disassembly, ma2011disassembly, kim2017optimal, lee2022task}, particularly mixed integer linear programming (MILP) \cite{ccil2022two, kucukkoc2020balancing, yin2022mixed, zhang2022improved, mutlu2021memetic}, are widely used to solve constrained optimization problem. Other optimization methods are also adopted for obtaining the optimal disassembly sequence, such as stochastic mixed-integer nonlinear programming \cite{behdad2014leveraging}, Tabu search \cite{alshibli2016disassembly, tao2018partial}, rule-based methods \cite{smith2016partial}, Q-learning \cite{chen2023disassembly}, multi-agent reinforcement learning \cite{xiao2023multi}, and teaching-learning-based optimization (TLBO) \cite{xia2013simplified}. The optimization methods and corresponding references are listed in Table \ref{tab:ch1_dsp_op}.

\begin{table}[!htbp]
	\centering
	\begin{tabular}{p{3.5cm} | p{4cm}}\toprule[1.5pt]
		\textbf{Optimization Method} & \textbf{References} \\\hline
		Genetic algorithms               & \cite{go2012genetically, kheder2015disassembly, kucukkoc2020balancing} \\\hline
		Artificial bee colony            & \cite{percoco2013preliminary, liu2018robotic, liu2020collaborative, wang2021discrete, kalayci2013artificial, ccil2022two} \\\hline
		Ant colony                       & \cite{luo2016integrated, malik2010performance} \\\hline
		Particle swarm                   & \cite{zhong2011disassembly, yeh2012simplified, kalayci2013particle} \\\hline
		Scatter search                   & \cite{li2013selective, kuo2013waste, guo2019lexicographic, guo2017dual} \\\hline
		Linear programming               & \cite{zhu2013disassembly, ma2011disassembly, kim2017optimal, lee2022task, ccil2022two, kucukkoc2020balancing, yin2022mixed, zhang2022improved, mutlu2021memetic} \\\hline
		Other modeling methods           &  \cite{behdad2014leveraging,alshibli2016disassembly, tao2018partial, smith2016partial, xia2013simplified} \\
		\toprule[1.5pt]
	\end{tabular} 
	\caption{Disassembly Optimization Methods} 
	\label{tab:ch1_dsp_op}
\end{table}

In our analysis of papers published between 2010 and 2022, we observed that artificial bee colony, ant colony, and linear programming emerged as the most widely used methods for disassembly optimization. The primary focus of these studies was to either minimize calculation time or enhance the quality of optimization outcomes. However, it is important to note that the applicability of these findings in real-world scenarios may be limited due to uncertainties inherent in the disassembly process. Consequently, there arises a pressing need to develop a disassembly sequence planner capable of dynamically re-planning the disassembly order and re-assigning workers in real-time, taking into account the actual status of disassembly tasks.

When discussing task allocation and sequence planning for HRC, one notable challenge in industries is the assignment of multiple human operators and robots to a sequence of assembly/disassembly tasks. This challenge arises due to the limited flexibility of robots and the programming efforts required for dynamic task assignment \cite{ranz2017capability}. Numerous publications have focused on addressing this optimization problem \cite{zanchettin2018prediction}. For instance, Wilcox et al. \cite{wilcox2012optimization} developed an adaptive preference algorithm (APA) to achieve optimal task assignments in human-robot teams. Chen et al. \cite{chen2013optimal} proposed a genetic algorithm for an HRC assembly workstation that aimed to minimize assembly costs, including time. Wu et al. \cite{wu2017toward} and Rahman et al. \cite{rahman2015trust} presented trust-based dynamic task allocation strategies to optimize assembly sequences.

A common approach for task allocation between human operators and robots is to determine an optimal assembly sequence that takes into account the capabilities of each individual \cite{ranz2017capability, johannsmeier2016hierarchical, tsarouchi2017human, schroter2016methodology, chen2013optimal}. Rosenfeld et al. \cite{rosenfeld2016human} implemented a robotic warehouse system with a human operator and multiple robots using the Advice Optimization Problem (AOP), which aims to maximize workers' performance. Subsequent work has focused on computationally driven approaches to automatically find the optimal assembly sequence  \cite{thomas2018learning}. Similarly, real-time task planning in HRC assembly can significantly contribute to achieving one or multiple objectives while considering constraints such as available workers (human operators/robots) and time consumption \cite{yu2020mastering}. However, limited research has been conducted at the task planning level to enhance HRC efficiency with task scheduling.

Dynamic sequence planning has been integrated into various algorithms to adapt and optimize real-time resource allocation. For instance, Lee et al. \cite{lee2020real} incorporated real-time human motion into the disassembly cost and formulated the HRC-DSP problem as a receding horizon optimization problem. Riedelbauch et al. \cite{riedelbauch2019exploiting} proposed an action selection algorithm that enables robots to dynamically choose pick-and-place operations that contribute to a shared goal with human operators. Moreover, the states of HRC, such as the fatigue level of humans and the positions/moving directions of robots and human operators, can be taken into account in assembly/disassembly sequence planning \cite{li2019sequence, aliev2019task, lee2020disassembly}. These factors contribute to a more comprehensive and efficient planning process in HRC.

The field of HRC manufacturing/remanufacturing has generated a substantial amount of literature, particularly concerning assembly lines and warehouse applications. Optimization methods are commonly employed to address the task allocation problem in HRC. These methods include trust-based optimal task allocation, Advice Optimization Problem (AOP) formulation, and optimized scheduling using integer linear programming. Some approaches parameterize the capabilities of human operators and robots to achieve optimal task distribution. The optimization efforts primarily revolve around two key aspects that form the basis of disassembly process planning: (1) disassembly sequence planning and (2) task allocation. These areas receive significant attention in the literature as they play crucial roles in optimizing HRC operations.

\textbf{Sequence planning.} The optimal task sequence is a critical challenge in DSP as it directly impacts the overall effectiveness of the disassembling process. Various challenges arise, such as allocating disassembly tasks in the correct order and selecting components efficiently from a large number of sub-assemblies within a limited time frame. Additionally, as the number of components to be disassembled increases, the number of potential solutions grows exponentially. The complexity of the entire disassembly process is further compounded by the need to select the appropriate worker from options including robots, human operators, or human-robot teams. Moreover, due to computational constraints, only a near-optimal solution can often be obtained, rather than an exact optimal solution. Furthermore, the presence of uncertainty in remanufacturing systems adds an additional layer of complexity to HRC planning. The planning process needs to account for uncertainties in the volume and condition of devices, as well as the inherent uncertainty associated with operations. Future HRC systems are expected to incorporate uncertainty considerations into their planning operations.  For example, Ye et al. \cite{ye2022self} introduced the concept of fuzzification in DSP, enabling disassembly sequence planning to adapt to failures and dynamically re-plan in real time. They proposed a dual-loop self-evolving framework that handles uncertain interference conditions, allowing for more robust and adaptive HRC planning.

\textbf{Task allocation.} Task assignment in collaborative systems plays a vital role in maximizing resource utilization and enhancing collaboration efficiency. Optimal and intelligent task assignments are crucial for achieving these goals. In collaborative disassembly workstations, task reassignment may occur based on factors such as the suitability of a disassembly tool for human operators or robots, their availability, and cost considerations. With the introduction of HRC, a disassembly task can be assigned to a robot, a human operator, or both simultaneously. As remanufacturing systems continue to grow in complexity, more advanced task-planning algorithms will be required. For example, Wurster et al. \cite{wurster2022modelling} developed a reinforcement learning approach for order dispatching control tasks involving humans and robots. Moreover, future HRC systems are expected to involve multiple robots and humans working together. Galina et al. \cite{galina2023approach} demonstrated how a mixed team of robots can collaborate to complete tasks and proposed an algorithm for task assignment based on factors such as availability and efficiency. Their algorithm represents the process as a graph, with events and operations as vertices connected by weighted edges representing time. These advancements in task-planning algorithms facilitate more effective coordination and utilization of resources in complex HRC systems.

\section{Bringing HRC to Disassembly: Robotic Motion Planning and Control (Technology)} 

Although robotic technologies have made significant advancements in the field of manufacturing over the past few decades, their applications in disassembly and remanufacturing remain quite limited. In traditional manufacturing settings, robots are typically confined within cages for safety reasons and are pre-programmed to perform repetitive tasks in controlled environments. For instance, prominent industrial robot company FANUC has supplied a large number of robots for applications such as welding, dispensing, sealing, material removal, and painting. More recently, with the growing interest in flexible automation, several collaborative robots (cobots) have been developed and brought to market. However, existing cobots often rely on rudimentary protection mechanisms, leading to a considerable loss in efficiency.

Disassembly processes, particularly in the context of remanufacturing, require robots to operate outside the confines of cages and work extensively alongside human operators, presenting significant new challenges for robotics. In fact, robotics technology on the remanufacturing side is still in its nascent stages of development. In this section, we aim to address two key aspects: (1) reviewing studies that focus on collision-free robotic motion planning to establish a safe shared workspace for human-robot collaboration, and (2) discussing the associated challenges specific to disassembly. We approach safe motion planning in human-robot collaboration by considering two critical steps: human motion prediction and robotic motion planning.

\subsection{Human motion prediction}

Due to the significant level of uncertainty involved in the disassembly process of end-of-use products, human operators often find themselves needing to adjust preplanned task sequences or undertake unplanned tasks based on the actual condition of the products being recycled. Such adjustments and additional tasks are typically not foreseen by the robot system in advance. In order to ensure the safety of human workers and enable seamless collaboration between humans and robots during disassembly tasks, cobots must possess the ability to comprehend human behavior and adapt their motion plans in real time. In this context, human motion forecasting plays a crucial role as an indispensable component of a safe and efficient HRC system. Numerous studies have been conducted in the field of computer vision to address human motion prediction \cite{fragkiadaki2015recurrent, martinez2017human, aksan2021spatio, alahi2016social, gupta2018social, wang2021multi, yuan2020dlow, mao2021generating, ma2022multi}. These studies have approached the topic from various research perspectives, such as accurate and deterministic motion prediction \cite{fragkiadaki2015recurrent, martinez2017human, aksan2021spatio}, prediction of multiple agents with interactions \cite{alahi2016social, gupta2018social, wang2021multi}, and stochastic motion prediction \cite{yuan2020dlow, mao2021generating, ma2022multi}. Despite the existence of such research works, the effective implementation and application of human behavior prediction in HRC are still being explored.

Probabilistic models have long been employed in the prediction of human motion, dating back to the early stage of research in this field. For instance, \cite{ding2011human} proposed a method based on the Hidden Markov Models (HMMs) to predict potential areas within the workspace that may be occupied by the human arm over a specific prediction horizon. This predicted region can subsequently used as a safety constraint in robot motion planning. Similarly, Mainprice and Otte \cite{mainprice2013human} computed workspace occupancy in real-time using Gaussian Mixture Models (GMMs) to encode a library of human motion. They constructed a probabilistic representation of workspace occupancy through Gaussian Mixture Regression (GMR) during the offline stage. In the online stage, observed trajectories were matched with the GMMs, and GMR was employed to extract the best-fitting motion. Luo et al. \cite{luo2018unsupervised} introduced a two-layer framework of GMMs, employing the features of palm position and arm joint center positions respectively. This two-layer structure demonstrated improved recognition performance while effectively modeling the entire arm trajectory. Additionally, they utilized an unsupervised online learning algorithm to update the models with newly observed trajectories, enabling adaptation to new agents and motion styles. These probabilistic methods are well-suited for capturing the stochastic nature of human motion but may face challenges when dealing with complex motion patterns.

In addition to the probabilistic model, Inverse Optimal Control (IOC) represents another promising approach for Human motion prediction in HRC. IOC-based methods assume that human motion is optimal with respect to an unknown cost function, which is typically defined as a linear combination of user-defined features related to motion trajectories for simplicity.  Mainprice and Otte \cite{mainprice2015predicting} utilized IOC to learn the underlying cost function from human-demonstrated trajectories in the scenario of human-human collaboration. They then predicted the motion of an active human through iterative motion re-planning based on the learned cost function. In a similar vein, Mainprice and Otte \cite{mainprice2016goal} introduced the concept of a Goal Set to relax the constraint of knowing the goal configuration in advance. Their work encompassed both human-human collaboration scenarios and human-robot workspace-sharing experiments. IOC was employed to capture human behavior in an HRC scenario in \cite{tian2023optimization}. In contrast to \cite{mainprice2015predicting, mainprice2016goal}, the cost function learning stage considered the collision cost as a feature instead of solely penalizing the distance between humans and robots during iterative re-planning. Moreover, additional constraints on the weighting vector were incorporated to prevent the algorithm from overly emphasizing specific features.

Deep learning techniques have recently made their way into HRC to capture the complex motion patterns exhibited by humans. Cheng et al. \cite{cheng2019human} proposed Semi-Adaptable Feedforward Neural Networks, which adapt the parameters of the last layer to accommodate uncertainties arising from time-varying human behavior and individual differences among human agents.  This approach was further combined with a model-based method in \cite{landi2019prediction} to predict human hand trajectories and final targets. Wang et al. \cite{wang2017collision} utilized a CNN-RNN-based model to predict human hand movements. Convolutional Neural Networks (CNNs) were employed to extract visual features from image inputs, while Recurrent Neural Networks (RNNs) predicted hand movements at the pixel level. Zhang et al. \cite{zhang2020recurrent} argued that standard RNNs might be ineffective for human motion prediction due to interactions among different body parts. They introduced component and coordination functional units into the RNN structure to analyze the evolutionary motion pattern of specific body parts and the coordination among different body parts. Additionally, Monte-Carlo dropout was investigated to improve the reliability of prediction results. To enhance prediction performance, \cite{liu2020human} \cite{liu2022dynamic} incorporated kinematic and dynamic information of human motion into neural networks. Liu et al. \cite{liu2020human} employed RNNs to predict human wrist motion, extending the prediction to full-arm motion using Inverse Kinematics. A modified Kalman filter was utilized to adapt the model in real time to different users or tasks. Liu et al. \cite{liu2022dynamic} considered the effect of muscle force on motion by incorporating a dynamic model informed by Lagrangian mechanics. The future muscle force was predicted using a neural network, and an unscented Kalman filter was used to handle the nonlinear arm dynamic model and generate future motion. Eltouny et al. \cite{eltouny2023tgn} used deep ensembles to predict human motion with uncertainty awareness. Liao et al. \cite{liao2022human} uses a combination of convolutional long short-term memory (ConvLSTM) and You Only Look Once (YOLO) to particularly predict human operations in disassembly tasks. Generative models have also been explored for human motion prediction in HRC. Butepage et al. \cite{butepage2017anticipating} trained a conditional variational auto-encoder (CVAE) using RGB depth images, enabling the generation of multiple future predictions based on the observed context. Tian et al. \cite{tian2023transfusion} proposed a practical and effective transformer-based diffusion model for 3D human motion prediction.

In addition to data-driven methods, certain researchers have directed their attention towards model-based approaches for human motion prediction. One such example is the integration of a minimum-jerk model-based algorithm for human motion prediction in \cite{dinh2015approach}, specifically designed to facilitate local obstacle avoidance in close HRC scenarios. Another approach, presented in \cite{oguz2017hybrid}, suggested a two-stage prediction method that combines the classical minimum-jerk model with Dynamic Movement Primitives (DMPs) to forecast human motion while taking into account obstacles present in the environment.

Besides 3D human pose prediction, another active research area in Human-Robot Collaboration (HRC) for assembly or disassembly tasks is the prediction of human actions or intentions. Similar to 3D pose prediction, there is a wide range of methodologies employed for human action prediction. For instance, in \cite{luo2019human}, Probabilistic Dynamic Movement Primitive (PDMP) was utilized to predict both human intention and hand motion. A Gaussian process-based method is proposed in \cite{li2020data} to infer human intentions in reaching tasks. In the case of assembly tasks, \cite{liu2017human} applied Hidden Markov Models (HMM) to model human motions as a sequence. In \cite{perez2015fast}, a Bayesian approach utilizing probabilistic flow tubes was employed to classify time series data and identify the current motion class being executed by a human. Several studies, such as \cite{liu2019deep, wang2017recurrent, wang2018deep}, investigated the use of CNN-RNN networks and Deep CNN to identify human intent from visual inputs. Additionally, \cite{petkovic2022human} proposed an action prediction method based on motion cues and gaze, using shared-weight long short-term memory (LSTM) networks and feature dimensionality reduction. Although these research works employ various methods, their common objective is to discern human actions or intentions based on historical observations.

Despite extensive research conducted in this field, incorporating human motion prediction into disassembly tasks still presents formidable challenges. Firstly, human motion itself is inherently complex. The highly nonlinear and time-varying nature of human movement makes it difficult to achieve accurate predictions. Secondly, human motion involves inherent uncertainty. Future motions of human workers can be influenced by various factors, such as unexpected product conditions, sudden changes in the environment, and interactions with other moving agents in shared workspaces. Reliable prediction results require sophisticated techniques that are currently lacking. 

Considering the complexity and uncertainty nature of human motions, the emerging deep learning techniques may bring some new opportunities in human motion prediction. However, the absence of widely-accepted real-world datasets in disassembly scenarios poses a significant challenge. Currently available benchmarks of human motion datasets, such as Human3.6M \cite{ionescu2013human3}, HumanEva-I \cite{sigal2010humaneva} and CMU Mocap \cite{Mocap}, provide various data formats, including sequential pose data, image data, and video data, and give researchers the flexibility to design their algorithm. Even though these datasets are widely used in human motion prediction, they are difficult to be directly applied to human modeling in the HRC problem. First, these datasets mainly focus on daily scenarios, such as walking, sitting, eating, and discussing, and are not specifically designed for HRC. Furthermore, the existing datasets only provide a limited amount of human motions with interactions with other people. These interactions are crucial in HRC, as one agent's motion will affect the other's motion. For example, the human and the robot will avoid collision with each other while performing different tasks in the shared space. And they will approach each other and coordinate their movements when they are jointly performing a task, such as lifting a heavy component or passing tools. Therefore, the dataset for HRC should consider both the human operator and the robot at the same time to capture such interactions. Moreover, to analyze and predict human motion more reliably in disassembly, it should involve the information of the end-of-used products which would affect the human intentions as well as the human-robot collaboration mode in real-time.
Considering the gap between benchmark datasets and the research question in collaborative electronic disassembly, we need to expand our current human motion dataset to one specifically designed for HRC disassembly. It is worth noting that, 
acquiring such high-quality HRC disassembly data for training and validation purposes is both expensive and time-consuming. Addressing these challenges and developing effective solutions remains an ongoing task in the field.

\subsection{Motion planning balancing between efficiency and safety}

The core challenge in robotic motion planning is to find a sequence of collision-free movements for a robot to reach a specific goal. Prior research has explored several methods to address this problem, categorized into different approaches such as grid-based, artificial potential fields, sampling-based, optimization-based, and learning-based methods. In this subsection, we will first examine the existing studies in this field and subsequently discuss the specific challenges associated with robotic motion planning within the context of collaborative disassembling processes.

Grid-based methods, introduced by Hart et al. \cite{hart1968formal}, involve discretizing the planning space into a grid and employing heuristic functions to estimate the cost of traversing different cells. These methods show promise in finding optimal solutions if they exist. However, a significant drawback arises when the dimension of the planning space increases, as this leads to a substantial rise in computational costs \cite{tang2012review}. Consequently, for collaborative disassembling processes involving robots with a high degree of freedom, grid-based methods may not be applicable. In addition to the inherent challenge of high-dimensional planning spaces, the presence of human motion in HRC further complicates the planning process \cite{lasota2017survey}. Human motion introduces additional complexity to the environment, continuously altering the available free planning space in real time. This dynamic nature necessitates planning algorithms that can swiftly adapt and respond to evolving human motion.

Sampling-based methods, such as Rapidly Exploring Random Trees (RRT) \cite{lavalle1998rapidly} and Probabilistic Roadmaps (PRM) \cite{kavraki1996probabilistic}, are effective in handling high-dimensional planning problems in narrow passages and offer probabilistic completeness. These methods randomly generate samples in the configuration space and connect feasible configurations to create a complete trajectory for the robot. In the context of collaborative tasks, \cite{rajendran2021human} proposed a human-aware RRT-Connect planner to generate manipulator motions that prioritize human safety. Additionally, \cite{wei2018method} introduced target directional node extension to enhance the sampling speed of the RRT algorithm, enabling the robot to dynamically respond to human motion. Furthermore, asymptotically optimal sampling-based methods like RRT* and PRM* have been developed to not only generate feasible robot motions but also optimize them. The optimality of the motions produced by these methods improves as the planning time increases. To enhance planning efficiency, recent approaches such as Batch Informed Trees \cite{gammell2015batch} and Fast Marching Trees \cite{janson2015fast} have been proposed. These methods enable quick identification of collision-free motions for high-degree-of-freedom manipulators while maintaining optimality.

In collaborative disassembling processes, robot motions need to fulfill collision-free requirements while also adhering to specific task constraints. These constraints may involve maintaining a particular orientation to facilitate fastening component disassembly \cite{stilman2007task} or following a desired path while carrying a disassembled component \cite{liu2023task}. However, sampling-based methods often struggle to provide real-time responsiveness in such scenarios \cite{kingston2018sampling}. To address these task constraints in disassembling processes, the motion planning problems for robots can be formulated and solved as optimization-based problems, inherently generating task-constrained robot motions. Examples of such methods include CHOMP \cite{ratliff2009chomp}, TrajOpt \cite{schulman2014motion}, and STOMP \cite{kalakrishnan2011stomp}. For instance, the work in \cite{zhao2020contact} solved an iterative convex optimization problem to generate contact-rich robot motions in a welding process. Another approach, presented in \cite{faroni2019mpc}, introduced a model predictive control framework that allowed the robot to slow down the task and maximize the distance from the operators when faced with close proximity.

In recent years, researchers have increasingly turned to neural networks to plan robot motions. These neural planners leverage the power of machine learning and expert demonstrations to swiftly generate collision-free robot motions. For instance, in \cite{wang2020neural}, a neural RRT* algorithm was proposed that predicts the probability distribution of optimal paths for given tasks. The work by \cite{khan2020graph} employed a graph neural network to identify critical sampling points in the configuration space, significantly speeding up the planning process. Furthermore, \cite{yu2021reducing} utilized a well-trained network to reduce collision-checking during path exploration and smoothing, enhancing overall planning efficiency. These approaches harness neural networks to improve specific modules of classical planners. Moreover, neural networks can also be used as complete planner pipelines. For instance, \cite{bency2019neural} employed recurrent neural networks to generate end-to-end robot motions iteratively. In another study, \cite{qureshi2020motion} presented a learning-based neural planner that considers the planning environment and generates a collision-free path connecting given start and goal configurations for the robot.

Integrating human motion prediction into robotic motion planning enables the robot to anticipate and respond to human actions, resulting in improved safety and efficiency in HRC scenarios. Several studies have demonstrated the benefits of incorporating human motion prediction into robot planning. For instance, in \cite{kanazawa2019adaptive}, the robot motion planning is adapted to the predicted human motion, effectively avoiding potential collisions and eliminating unnecessary waiting time in collaborative tasks. Similarly, in \cite{cheng2020towards}, both task plan recognition and human trajectory prediction are utilized for robot planning, resulting in a significant reduction in task execution time. In \cite{park2019planner}, the predicted human motion is used to establish collision boundaries, allowing safe manipulator motions to be computed within these boundaries. This approach ensures the robot avoids collisions while performing its desired tasks. In \cite{liu2023task}, the predicted human motion, along with associated uncertainty from the prediction model, is converted into dangerous zones for safe manipulator planning. This enables the robot to proactively avoid human workers while simultaneously fulfilling its tasks. In brief, extensive robotic motion planners that incorporate human motion prediction have been developed in recent years, with a primary focus on ensuring human safety in HRC applications. These approaches demonstrate the importance of considering human motion in robot planning to enhance safety and efficiency. 

Despite significant progress in developing robotic planning approaches for collaborative disassembling processes, there is no one-size-fits-all solution. Planning in such scenarios requires a delicate balance between efficiency and safety. However, several challenges persist in this domain. Firstly, disassembly tasks encompass a wide range of complexity levels. Planning efficient robot motions to accomplish disassembling tasks involves navigating high-dimensional spaces while considering task-specific constraints, dynamic constraints, and additional objective requirements. This complexity adds to the difficulty of finding optimal solutions. Secondly, the involvement of human workers further complicates the planning process. Safety concerns necessitate real-time responsiveness from the robot. It must plan collision-free motions in a continuously changing planning space and execute these motions in real-time. Moreover, human workers' movements can be unpredictable, introducing uncertainty that the planning algorithm must effectively handle to ensure human safety. In summary, the planning of collaborative disassembling processes remains a significant challenge that requires ongoing efforts and attention. Balancing efficiency and safety, handling the complexity of disassembly tasks, and addressing the uncertainties introduced by human movement are vital aspects that need to be continually addressed and improved upon.

\section{Bringing HRC to Disassembly: Human factors and human-robotics system integration (Workers)}

Workplace risks associated with HRC in disassembly can be substantial. Even survey data on robot-related injuries may not fully reflect such risks, partially due to the fact that HRC applications in this field have only emerged in recent years. The implementation of collaborative human-robot systems in shared spaces would expose workers to even higher risks of injury or death if associated safety research is not emphasized \cite{malm2010safety}. Therefore, ensuring human safety is a prerequisite for any robotics application. A few standards have taken operational safety during HRC into consideration. For example, ISO 10218 is the robots and robotic devices safety package. However, it emphasizes more on the manufacturing requirements for robots and does not address personnel safety \cite{fryman2012safety}. ISO 15066 standard supplements the above standard on limited collaborative industrial robot operations. In brief, substantial knowledge gaps in occupational safety regulations and collaborative robot application need to be bridged before humans and robots can routinely, safely, and comfortably share the same environments \cite{goodrich2008human, sheridan2016human}. In addition, merely avoiding physical contact is insufficient in ensuring safety, as harm could occur to a person from other venues (e.g., the excessive mental workload could have substantial negative impacts). Even though the importance of safety and its applications have been well acknowledged, research studies leveraging system-level objective methods are still rare. Furthermore, multiple variables may significantly influence the environment and safety, such as stress, situation awareness and risk perception during operations, implementation of varying levels of automation, and operator acceptance of emerging safety interventions. All of these variables need to be considered in safety measurement processes.

There are many human factors, such as human safety, discomfortability, fatigue, human ergonomics, and mental stress, that need to be considered. For example, as the human operator works closely with the robot, the human worker's discomfort may increase \cite{lasota2013developing} and there could be high risks of collision between human operators and robots. In the domain of human-robot interaction (HRI) \cite{cherubini2016collaborative, lasota2017survey} and remanufacturing \cite{daneshmand2022industry}, human safety has been considered in the design and operation of a collaborative work cell, including the prevention of accidents due to unexpected robot motions \cite{robla2017working} and health problems from hazardous work environments \cite{gerbers2018safe}. Comparing the human operator's capability to a robotic system, the performance of manual assemblies is affected by safety considerations and ergonomic factors \cite{matheson2019human}. For instance, lifting heavy items could lead to back injuries, and the product quality could fluctuate due to human fatigue \cite{li2019sequence, zhang2022cycle}. 
Studies on human factors including ergonomics have shown the impact of stress on work performance and human-machine interactions \cite{proctor2018human}. 
The objective of ergonomics intervention is often to avoid musculoskeletal disorders\cite{baykasoglu2017modeling, xu2012design}, which could be the result of handling forceful, long-term, monotonous and repetitive tasks \cite{li2019sequence, arkouli2021ai}. Most existing techniques for human ergonomics are based on offline processing of human motions and the work environments, and thus a fast reconfigurable HRC workstation was developed to incorporate human factors in real-time to ensure productivity \cite{kim2019adaptable}. Meanwhile, studies have been carried out to parameterize the ergonomics using risk assessment methodology. In \cite{maurice2017human}, an ergonomic assessment method was presented to optimize robot actions in collaborative tasks. \cite{faber2016model} offered a proposal to obtain the optimal HRC assembly sequence with the model of ergonomic risk assessment.
The occupational repetitive action (OCRA) method \cite{gualtieri2020design}, known as a checklist approach was also used to present the risk assessment model. For example, in \cite{baykasoglu2017modeling}, ergonomic risks were considered as constraints using OCRA to minimize the number of workstations. And in \cite{botti2017integrating}, OCRA was used to evaluate the ergonomic risk levels with an integer linear programming model to design HRC assembly lines. Moreover, the assembly system design \cite{baykasoglu2017modeling, ranz2017capability, malik2019complexity} was often used to assess the level of ergonomic risk. For instance, \cite{tram2020optimal} proposed a time-cost optimization model to evaluate the ergonomics difficulty index via a score sheet in the assembly system design. A different approach using the ``mental model" was introduced in \cite{akkaladevi2016human, tabrez2020survey, nikolaidis2012human} taking into account human preferences, knowledge of the tasks, and the capabilities of the human operator and the robot to complete tasks safely and efficiently. In short, to reduce assembly/disassembly costs, human factors including ergonomics should be taken into account to improve productivity and ensure the safety and mental health of human operators.

The paucity of research in human factor analyses in disassembly environments, notwithstanding its critical significance, underscores the need for focused exploration in this area. 
It's worth mentioning that other sections of our paper, such as those dealing with human motion prediction or robot path planning, do contain references to papers that touch upon elements of human factors. These were not included in this section because their primary focus did not revolve around HF/E topics.
The limited number of application-based experimental studies currently available provide a rudimentary understanding of the implications of HRC for designing safe and efficient human-centric systems \cite{cheng2020towards,chen2022human}. A key constituent of these analyses is the biomechanical ergonomic studies that explore physical safety considerations. These studies are crucial in identifying risk factors contributing to the potential onset of long-term work-related musculoskeletal disorders \cite{wang2019usability}. In an investigation aimed at understanding the impact of robot collaboration on hardware component extraction from desktops in an e-waste disassembly scenario, a noticeable decrease in musculoskeletal effort was observed among human participants when working with the robot \cite{cheng2020towards}. In this particular collaborative task, there was a demonstrable reduction in the propensity towards lower back pain and shoulder disorders, which highlights the crucial role of biomechanical ergonomics in HRC.
Alongside physical safety, human factor analyses also account for psychological factors such as cognitive workload, which can have a profound influence on task performance and muscle activity \cite{biondi2021overloaded}. An apt evaluation of cognitive states is thus an essential component of the comprehensive safety protocol in human-machine interactions, particularly in disassembly tasks. In a study evaluating the efficiency of workers using an augmented reality system (Google Glass) for task guidance during phone disassembly, it was found that while error avoidance was enhanced, the mental workload concurrently escalated, thereby reducing the overall performance of the augmented reality system \cite{wang2019usability}.
While the application environment e-waste disassembly we discuss here is unique, HF/E analysis can often draw upon a rich body of knowledge derived from analogous studies conducted in more generalized occupational settings. For instance, the assembly processes explored in our work are quite common in traditional manufacturing settings, and they can often be seen as the reverse of disassembly processes in many instances.
Similarly, many relevant HRC standards and policies can be adapted to the disassembly environment. These standards and regulations typically apply broadly across industries and are not confined to specific sectors. However, we acknowledge that as the re-manufacturing sector continues to expand and more HRC approaches are deployed within this domain, future safety standards and robotics regulations may incorporate specific provisions for this area. Particularly, there are certain aspects of environmental factors such as high uncertainty, less structural of the working environment, that make the disassembly unique and needs special attention (e.g., new safety standard for this business sector). This could represent a significant research avenue, not only for the HF/E community but also for other disciplines such as safety engineering, robotics, and control, among others.

In summary, while the potential benefits of HRC in disassembly are evident in reduced physical stress and enhanced task accuracy, further research is needed to identify an optimal balance between physical ease and cognitive load. This pursuit of balance will pave the way for designing human-centric systems that promote both physical and mental well-being during disassembly operations.

\section{Bringing HRC to Disassembly: Economic Considerations and Implications for Human Workforce (Work)}

While HRC-enabled disassembly promises efficiency enhancement and cost savings, several economic aspects and labor force implications demand thoughtful consideration. 

\textbf{Economic Feasibility of HRC in Disassembly.} Remanufacturers' decision to incorporate Human-Robot Collaboration (HRC) in disassembly processes is primarily driven by economic costs and benefits. This involves weighing the initial investment in robotics, their maintenance, and associated operational costs against the potential productivity gains and efficiencies they could bring \cite{matheson2019human}. HRC offers an opportunity for faster, more precise disassembly, reducing waste and potentially enabling higher levels of component reuse \cite{parsa2021human, cohen2022deploying}. Continuous operation, reducing downtime typically associated with purely human-operated systems, is another potential benefit of HRC. Despite these advantages, the return on investment depends largely on the nature of the disassembly task, disassembly completion time, the cost of human labor, the level of complexity in the physical motion of the human operator and the robot, and the market value of reclaimed materials \cite{zhou2022stackelberg}. Some studies have proposed algorithms to maximize profit by altering several parameters of the HRC disassembly process. For example, \cite{xu2020human} proposed a method of integrating stochastic simulation with the artificial bee colony algorithm to ensure profit maximization by minimizing energy usage, difficulty level of disassembly operations, and the number of workstations. An automated unfastening method was proposed in \cite{li2020unfastening} to remove screws from a turbocharger cost-efficiently. There are other suggestions in the literature for maximizing profits with disassembly sequencing planning \cite{liu2021optimizing, lu2020hybrid, jin2017systematic}; while these studies do not explicitly consider HRC, the proposed profit-maximizing plans could be adapted to the case of disassembly with HRC.

A significant barrier to the wider adaption of HRC-enabled disassembly is that HRC workstation designs and disassembly procedures are typically customized for specific products. In other words, there are no standardized systems that are applicable to a wide range of disassembly scenarios. Low-volume and high-mixture of EoU products could make it difficult to implement HRC, especially among small- and medium-sized enterprises (SMEs), as these enterprises typically lack the capital needed to incur large initial costs of setting up collaborative robots \cite{kim2019adaptable, gerbers2018safe}. Therefore, another opportunity for further advancement and adaption of HRC among remanufacturers is the development of cost-effective workstations that allow human operators and robots to handle different disassembly tasks and variants of EoU products.

\textbf{Implications for Human Labor Displacement and the Need for Upskilling.} The rise of HRC has sparked concerns over potential labor displacement. Automation in general may lead to job losses, particularly for those involved in manual disassembly tasks, exacerbating socio-economic disparities \cite{sen2022impact}. On the other hand, HRC also presents opportunities for upskilling human workers and facilitating job transformation. It can allow workers to move away from repetitive, dangerous, or labor-intensive tasks, redirecting human labor towards more strategic, creative, and value-added functions \cite{acemoglu2018race, acemoglu2019automation}. The evolution of HRC could help shape a future where humans and robots work together, complementing each other's capabilities \cite{willcocks2020robo}.

In addition, the demand for professionals who are adept at managing and maintaining these new robotic systems will increase over time, fostering new career paths. Proper training programs and educational policies should be in place to ensure that the human workforce is equipped to adapt to these shifts in the workplace. Upskilling workers to work alongside robots, maintaining and repairing robot systems, and managing HRC workflows can require substantial investment in training and education, which is a significant challenge facing the wider adoption of collaborative robots in disassembly.

\textbf{Balancing Technological Advancement with Socio-Economic Responsibility.} Under the lens of the circular economy, a profitable product disassembly with HRC could motivate the remanufacturers to collect more EoU products, reinforcing sustainability objectives. As we grapple with the urgent need to transition to a more circular economy, HRC in disassembly can play a crucial role in waste reduction, resource recovery, and extended product lifecycles. However, it is important that the economic benefits of this technology do not undermine social equity \cite{calvo2022evaluation}. Strategies need to be developed that simultaneously foster technological progress and socio-economic responsibility. This could involve stakeholder collaboration to enact inclusive labor policies, investment in skill development, and equitable access to new job opportunities, alongside a commitment to environmental sustainability. Thus, HRC must be positioned not just as a tool for operational efficiency, but as a mechanism to drive responsible and inclusive growth in the circular economy \cite{meier2023evaluation}.

An estimated total of 44.7 million metric tons of e-waste was generated globally in 2016, containing raw materials valued at the US \$64.7 billion \cite{balde2017global}. In the US alone, 2.37 million tons of e-waste (excluding appliances) were generated in 2016, and only about 25 percent of this waste (in weight) was recycled domestically \cite{report}. The answer to successful e-waste recovery lies in economics. While there are valid concerns about costs and challenges associated with the implementation of HRC in disassembly, the potential benefits and opportunities make a compelling case for this transition. At the core, HRC is about synergy—leveraging the strengths of both humans and robots to create value that exceeds what each can achieve independently. Future research questions that can facilitate a smooth transition to HRC-assisted smart disassembly include, a) under what conditions human-robot collaborative systems are economically feasible for U.S. remanufacturers? b) what are the estimated effects of broader adaptation of human-robot collaborative systems for aggregate employment, wages, and value-added in an economy? and, c) what economic and social policies are needed to better prepare society for a future that involves working in human-robot collaborative production systems? Transitioning to HRC in disassembly should be a thoughtful strategy to create shared values for businesses, workers, and society at large.

\section{Identified Key Challenges and Conclusions} \label{sec:rv2}  

The incorporation of HRC within remanufacturing facilities offers substantial sustainability benefits and enhances the quality of human work experiences. However, it also introduces a set of interconnected challenges that affect technology, workers, and the overall work processes. While HRC has been explored in the context of manufacturing, its specific implications in the realm of disassembly remain relatively underexplored. Particularly, in the case of EoU products characterized by traits like low-volume production, high diversity, and variable quality, the integration of collaborative disassembly presents challenges for both human operators tasked with efficient disassembly and robots striving for complete automation.  This is particularly relevant for SMEs that may possess limited resources for adaptation \cite{kim2019adaptable}. This section delves into the multifaceted challenges associated with the introduction of HRC into the disassembly process, drawing from our extensive review of recent literature.  We have considered not only papers directly related to HRC in disassembly but also research in related sub-domains.

Disassembly cannot be simply considered as the reverse of assembly: current manufacturing design guidelines, such as the minimization of part counts and the use of self-fastening components, tend to prioritize assembly efficiency at the expense of disassembly. Moreover, disassembly often involves a diverse array of EoU products, each with uncertain quantities, varying quality, and differing conditions.  Consequently, the implementation of HRC in disassembly processes must meet three critical criteria: (1) efficiency, (2) affordability, and (3) adaptability to the wide range of EoU products characterized by substantial uncertainties. The necessity for adaptability, from a technological standpoint, places a premium on precise human motion prediction—a formidable task given its complexity and nonlinearity inherent in disassembly activities.  This complexity is further compounded by the unpredictability introduced by environmental factors and interactions with other agents. Moreover, the absence of widely-accepted real-world datasets specific to disassembly poses a significant hurdle in training and validating prediction models. Efficiency and cost-efficiency pose another technological challenge in the implementation of HRC for disassembly, particularly when dealing with EoU products with limited resale value. Existing safety mechanisms, such as halting or slowing down robots when they come into proximity with human workers, can substantially impede overall efficiency in HRC scenarios, making the application of HRC in disassembly significantly more challenging. Thus, the paramount objective becomes the maximization of operational efficiency while minimizing cost, all the while ensuring the utmost safety for human workers. This balance is essential to facilitate the widespread adoption of HRC in disassembly processes. Conquering these multifaceted challenges and devising effective solutions remains an ongoing pursuit within this research domain.

When considering the perspective of workers in HRC scenarios for disassembly, it becomes evident that ensuring physical safety alone is insufficient.  This is because the disassembly environments for EoU products can often be characterized by poor structure, disorder, dirtiness, and potential hazards. Unlike assembly sites, which are typically clean, well-organized, and equipped with advanced technology, workers in disassembly settings face a higher vulnerability to excessive mental stress and psychological workload. Therefore, the design of HRC for disassembly must prioritize the mental and psychological well-being of workers to foster comfortable interaction and collaboration between humans and technology. Addressing these aspects introduces new challenges, particularly in the development of comprehensive and dependable measurement methods to evaluate the mental and psychological health of workers engaged in HRC disassembly. Moreover, from an occupational safety standpoint, existing standards such as ISO standard 10218 generally do not explicitly account for personnel safety \cite{fryman2012safety}, let alone in the context of disassembly environments. These complexities in safety considerations extend beyond those typically encountered in manufacturing sites. A noteworthy gap emerges in the realm of regulations and policies related to the mental and psychological health of human workers and occupational safety within the context of disassembly. These gaps must be effectively addressed before the widespread implementation of HRC in disassembly and re-manufacturing can realized. It is important to note that this particular aspect remains an underexplored area in existing literature, highlighting the need for further research and policy development in this crucial domain.

Regarding work, while substantial efforts have been made to model the effects of artificial intelligence on labor markets and the broader economy, recent studies \cite{berg2018should,acemoglu2018modeling} have yet to comprehensively account for the distinct characteristics of the emerging remanufacturing sector, particularly within the context of advanced HRC systems. Nevertheless, the potential for significantly enhanced efficiency and productivity through HRC in disassembly is undeniable. By entrusting robots with repetitive or hazardous tasks, human labor can pivot towards more strategic roles, thereby elevating the overall quality of work. The evolution of disassembly with HRC has the power to catalyze a transformation in the workforce, ushering in new job opportunities and roles. The future of disassembly labor within an HRC framework envisions a collaborative environment where humans and robots synergize to leverage their respective capabilities. However, realizing these benefits hinges on meticulous planning and strategic investment. It is crucial to recognize that there are tangible costs associated with the adoption of this technology. Expenses linked to technology acquisition, training, safety measures, and change management are very much real. For example, existing HRC-enabled workstations designed for specific products may not be economically adaptable to the disassembly of EoU products, given their unique characteristics of low-volume and high diversity. SMEs, which constitute a significant sector in the disassembly and recycling of EoU products, often grapple with limited capital resources required for the initial setup costs of collaborative robots. Thus, the challenge lies in finding ways to reduce the cost of HRC-enabled disassembly, encompassing technology acquisition, training, safety measures, and more, while simultaneously maximizing the profitability of recycling EoU products.
Nonetheless, when evaluating these investments in the broader context of potential benefits, they can be deemed as indispensable steps. Such investments not only hold the promise of economic prosperity but also aspire to shape a future where businesses operate at heightened productivity levels, jobs are safer and more fulfilling, and the manufacturing sector embraces sustainability as a guiding principle.

Beyond the interconnected challenges discussed earlier, the integration and implementation of HRC in disassembly introduce an even greater level of complexity. Several critical loops in this ecosystem remain insufficiently explored and unresolved: (1) Product Design Guidelines: Existing product design guidelines have yet to encompass considerations related to HRC-enabled disassembly. The current design principles tend to overlook the intricacies of disassembly with human-robot collaboration. (2) Occupational Safety Standards: Present occupational safety standards predominantly emphasize short-term physical safety, particularly the prevention of collisions and accidents. The broader aspects of safety, especially concerning mental and psychological well-being, remain less emphasized. (3) Economic Literature: Conventional economic literature has not sufficiently addressed the modeling of HRC production systems and their implications for labor markets. Often, these models assume a scenario where robots entirely replace human labor rather than focusing on how robots can complement human work. Additionally, these models may not adequately account for the unique dynamics of the remanufacturing sector, where data on economic activities are limited. The absence of sector-specific economic data in the remanufacturing domain further complicates the development of accurate models and analyses.

In summary, the complexities of incorporating HRC into disassembly processes are multifaceted, encompassing technology, workers, work processes, and their seamless integration. Although the potential advantages are significant, it is imperative to engage in meticulous planning, substantial investment, and effective coordination across various domains to successfully navigate these challenges. Such efforts are instrumental in shaping a future where HRC in disassembly translates into heightened productivity, increased job satisfaction, and a more sustainable approach to industrial processes.

\section*{Acknowledgment}
We would like to thank the following people who also assisted with the paper: Sibo Tian for the human motion prediction section, Wansong Liu for the robotic motion planning section, Haoyu Liao for the disassembly sequence planning section, Nicolas Grimaldi for the human factors section, and Fernando Brito for the economic implications section. 

This material is based upon work supported by the Future of Work at the Human-Technology Frontier (FW-HTF) Program of the National Science Foundation - USA under Grants No. 2026533 and No. 2026276. Any opinions, findings, conclusions, or recommendations expressed in this material are those of the authors and do not necessarily reflect the views of the National Science Foundation, the University at Buffalo, or the University of Florida. 

\normalem
\bibliography{reference}{}
\bibliographystyle{IEEEtran}
\end{document}